\documentclass{article}

\usepackage{arxiv}

\usepackage[utf8]{inputenc} 
\usepackage[T1]{fontenc}    
\usepackage{hyperref}       
\usepackage{url}            
\usepackage{booktabs}       
\usepackage{amsfonts}       
\usepackage{nicefrac}       
\usepackage{microtype}      
\usepackage{graphicx}
\usepackage{natbib}
\usepackage{doi}

\usepackage{amsmath}
\usepackage{amssymb}
\usepackage{mathtools}
\usepackage{amsthm}

\theoremstyle{plain}
\newtheorem{theorem}{Theorem}[section]
\newtheorem{proposition}[theorem]{Proposition}
\newtheorem{lemma}[theorem]{Lemma}

\theoremstyle{definition}

\theoremstyle{remark}

\usepackage{caption}

\newcommand{\bb}[1]{\mathbb{#1}}
\newcommand{\mbf}[1]{\mathbf{#1}}
\newcommand{\mc}[1]{\mathcal{#1}}

\newcommand{\Var}{\text{Var}}

\usepackage{ulem}
\usepackage{comment}

\usepackage{accents}

\usepackage{booktabs}
\usepackage{wrapfig}
\usepackage{sidecap}
\usepackage{nccmath}
\usepackage{dblfloatfix}

\makeatletter
\newcommand*\eqsize{%
  \@setfontsize\eqsize{8.9}{9.0}%
}
\makeatother

\makeatletter
\newcommand*\eqsmall{%
  \@setfontsize\eqsmall{6.7}{9.0}%
}
\makeatother

\NewDocumentEnvironment{FullWidth}{ +b }{
    \twocolumn[{#1}]%
}{}

\usepackage{comment}
\usepackage{booktabs}
\usepackage{tcolorbox}


\title{Data-Aware Random Feature Kernel for Transformers}

\author{Amirhossein Farzam\thanks{Correspondence: a.farzam@duke.edu}  \\
	Google DeepMind\\
	\And
	Hossein Mobahi \\
	Google DeepMind
    \And
	Nolan Andrew Miller \\
	Google DeepMind
    \And
	Luke Sernau \\
	Google DeepMind
}

\renewcommand{\shorttitle}{Data-Aware Random Feature Kernel for Transformers}

\hypersetup{
pdftitle={darkformer},
pdfsubject={darkformer},
pdfauthor={Amirhossein Farzam, Hossein Mobahi, Nolan Miller, Luke Sernau},
pdfkeywords={Efficient Transformers, Attention, Random Feature Kernels},
}

\begin{document}
\maketitle

\begin{abstract}

Transformers excel across domains, yet their quadratic attention complexity poses a barrier to scaling. 
Random-feature attention, as in \textit{Performers}, can reduce this cost to linear in the sequence length by approximating the softmax kernel with positive random features drawn from an isotropic distribution. 
In pretrained models, however, queries and keys are typically anisotropic. 
This induces high Monte Carlo variance in isotropic sampling schemes unless one retrains the model or uses a large feature budget.
Importance sampling can address this by adapting the sampling distribution to the input geometry, but complex data-dependent proposal distributions are often intractable.
We show that by data aligning the softmax kernel, we obtain an attention mechanism which can both admit a tractable minimal-variance proposal distribution for importance sampling, and exhibits better training stability.
Motivated by this finding, we introduce \textit{DARKFormer}, a \textbf{D}ata-\textbf{A}ware \textbf{R}andom-feature \textbf{K}ernel trans\textbf{former} that features a data-aligned kernel geometry.
DARKFormer learns the random-projection covariance, efficiently realizing an importance-sampled positive random-feature estimator for its data-aligned kernel.
Empirically, DARKFormer narrows the performance gap with exact softmax attention, particularly in finetuning regimes where pretrained representations are anisotropic. 
By combining random‑feature efficiency with data‑aware kernels, DARKFormer advances kernel‑based attention in resource‑constrained settings.
\end{abstract}

\section{Introduction}
\label{sect:intro}

The remarkable success of transformers is largely attributed to self-attention's ability to model long-range dependencies while flexibly adapting to their input.
Yet the quadratic complexity of attention mechanisms remains a critical limitation, especially for applications requiring long sequences.
Random feature kernel approximations, as in \textit{Performers}~\citep{choromanski2020rethinking}, address this by linearly approximating the softmax kernel in a feature space, achieving complexity that is linear in sequence length times the sample size. 
Under uniform sampling, this is an asymptotically unbiased estimator for the softmax kernel.
This uniform sampling strategy, however, leads to high Monte Carlo variance when inputs exhibit anisotropic structure.
To achieve low estimation error with anisotropically distributed queries and keys, as is common in real-world data~\citep{godey2024anisotropy}, the unbiased uniform estimator requires either large feature samples to reduce variance or extensive training to reshape input distributions toward isotropy.
Importance sampling offers a principled solution by adapting the sampling distribution to match data geometry, but implementing the optimal scheme depends on input-dependent matrices that are unknown a priori and involve non-trivial computations. 
Motivated by these challenges, we propose a \textbf{D}ata-\textbf{A}ware \textbf{R}andom-feature \textbf{K}ernel for trans\textbf{former}s---\textit{DARKFormer}---that learns a data-aligned kernel geometry and is capable of realizing importance sampling automatically. 
This improves the performance and stability while achieving the promised efficiency of random-feature-kernel-based attention.

Building on the random feature paradigm for efficient attention, DARKFormer replaces the dot product in standard softmax attention with a Mahalanobis inner product, obtaining a data-aware kernel that automatically corrects for anisotropy.
The random-feature estimator in DARKFormer learns a kernel geometry matrix $\Sigma$ (parameterized via $\Sigma=M^\top M$), yielding a linear-time approximation that implicitly implements an importance-sampling scheme and can yield lower-variance estimates.
Because the sampling strategy can adapt to the input distribution, queries and keys need not be isotropically distributed for DARKFormer to yield accurate approximations.
It naturally takes more draws where input density is enriched and fewer where it is depleted. 
We show that this is equivalent to importance sampling in the feature space, allowing DARKFormer to learn a data-aligned sampling geometry without explicitly computing per-sample importance weights. 
Consequently, DARKFormer enhances approximation under limited feature budgets, improving performance without requiring large samples. 
Furthermore, sampling random features according to input geometry can improve training stability, as we observe empirically.

DARKFormer is particularly suitable for finetuning regimes, where attention input distributions are dictated by pretrained weights, demanding substantial retraining to produce isotropic query-key distributions.
The learned covariance can be interpreted as learning a linear re-embedding.
When this covariance matches (or approximates) the inverse input covariance, this re-embedding can (approximately) whiten the queries and keys.
After this whitening, the attention kernel takes the standard softmax form.
Our analysis in sections~\ref{sec:variance} and \ref{sect:darkformer} motivates why aligning the sampling geometry with the representation statistics can reduce Monte Carlo error. 
We corroborate these findings through experiments on the Gemma model \citep{team2024gemma,team2024gemma2,team2025gemma3}.
Our empirical observations show that DARKFormer narrows the performance gap with exact attention, enjoys more stable training dynamics, and does not require a large sample, extensive retraining, or thorough hyperparameter tuning to achieve these improvements.
Notably, DARKFormer excels especially well in finetuning settings, requiring only a small amount of data to produce robust covariance matrices and adapt to the presence of whitening.
Together, these results show that this data-aware framework can achieve the linear complexity of random-feature attention with lower sampling costs via importance sampling, opening new possibilities for resource-efficient transformer architectures in applications where quadratic complexity remains prohibitive, such as long-context modeling and high-resolution vision tasks.

\begin{tcolorbox}[title=Main Contributions]
\textbf{Importance Sampling via DARKFormer.} 
We introduce \textit{DARKFormer}, which implements data-aligned random feature attention through a learnable covariance matrix, achieving low sample complexity with a tractable proposal distribution.

\vspace{.1cm}
\textbf{Variance Optimality and Data-Aligned Sampling.} 
We observe that variance-optimal random-feature estimators require data-aligned sampling. 

\vspace{.1cm}
\textbf{Efficiently Improves Performance.}
Experiments on Gemma demonstrate that DARKFormer narrows the performance gap with exact attention, with particular benefits in finetuning scenarios where query-key distributions are anisotropic.

\vspace{.1cm}
\textbf{Resource-constrained Finetuning.}
Our results show that DARKFormer can improve performance without large feature samples, long training cycles, or extensive hyperparameter tuning. Most notably we show that it is not necessary to train a model from scratch. The method is compatible with finetuning from pre-trained weights.
This makes it particularly suitable for resource-constrained environments.
\end{tcolorbox}

\section{Background}
\label{sect:background}

Exactly computing the softmax attention entails quadratic cost in sequence length, posing a fundamental challenge for scaling. 
This has motivated significant research in efficient approximations~\citep{choromanski2020rethinking,pengrandom,likhosherstov2023favor}, with random feature methods offering elegant  linear-complexity solutions.
We briefly review the attention mechanism and its random feature approximation that forms the foundation for our approach.

\paragraph{Attention Mechanism in Transformers.}
The fundamental building block of the Transformer architecture is the \textit{attention} mechanism, which captures pairwise interactions between tokens. 
Consider a sequence of $L$ input tokens, each with a $d$-dimensional embedding, mapped via learnable projection weights to 
\(
    Q, K, V \,\in\, \mathbb{R}^{L \times d},
\)
which are referred to as \textit{queries}, \textit{keys}, and \textit{values}, respectively. 
The classical scaled dot-product attention \cite{vaswani2017attention} is then defined by
\[
\mathrm{Att}(Q,K,V)
    =
    \mathrm{softmax}\!\Bigl(\tfrac{1}{\sqrt{d}}\,QK^\top\Bigr)\,V \;\in\;\mathbb{R}^{L \times d}.
\]
The softmax matrix is formed by entry-wise exponentiation of the dot-product kernel, where the $(i, j)$ entry captures the similarity between the $i^{\text{th}}$ query and the $j^{\text{th}}$ key, normalized to obtain a probability distribution over tokens for each query. 
This yields $L \times L$ attention weights, which, when multiplied by $V$, produces the output of the attention module.
The corresponding time and memory complexities are $\mc{O}(L^2\,d)$ and $\mathcal{O}(L^2)$. 
When $L$ is large, this becomes prohibitive in speed and storage. 
Various approximations and alternative formulations have therefore been proposed to mitigate this bottleneck.

\paragraph{Random Feature Approximation of Kernels.}
\citet{rahimi2007random} introduced random feature expansions as an efficient way to approximate kernel inner products.
Utilizing this foundational technique we can obtain efficient approximations of attention by replacing the softmax kernel $\kappa_{\text{SM}} (\mbf{q}, \mbf{k}) \coloneqq \exp(\mbf{q}^\top\mbf{k})$ by a random feature expansion
\footnote{For mapping this to $\text{Att}(Q,K,V)$, $\mbf{q}$ and $\mbf{k}$ in $\kappa_{\text{SM}}$ can be viewed as queries and keys that have absorbed the scaling by $1/\sqrt{d}$.
Normalizing this expression then produces softmax attention.}. 
Suppose we wish to approximate a kernel 
\(
\kappa(\mbf{x},\mbf{y}) 
\)
using an inner product of finite-dimensional feature maps,
\[
\kappa(\mbf{x},\mbf{y}) \approx \phi(\mbf{x})^\top \,\phi(\mbf{y}).
\]
We can construct such $\phi(\cdot)$ by sampling a set of random projections and suitably normalized basis functions. 
For instance, using trigonometric basis functions, we can construct the Gaussian kernel, $\kappa_{\text{Gauss}}(\mbf{x},\mbf{y}) = \exp\bigl(-\tfrac12\|\mbf{x}-\mbf{y}\|^2\bigr)$, by drawing $m$ samples $\omega_1,\dots,\omega_m \sim \mc{N}(0, I_d)$, defining
\begin{align*}
    \phi_{\mbf{\omega}}^{\text{trig}}(\mbf{x}) 
    =
    \tfrac{h(\mbf{x})}{\sqrt{m}}\,
    \Bigl[
    &\cos
    \bigl(
    \omega_1^\top \mbf{x}
    \bigr), 
    \;\dots,\;\cos
    \bigl(
    \omega_m^\top \mbf{x}
    \bigr), 
    \;
    \\
    &\sin \bigl(
    \omega_1^\top \mbf{x}
    \bigr),
    \;\dots,
    \;\sin \bigl(
    \omega_m^\top \mbf{x}
    \bigr)
    \Bigr]^\top,
\end{align*}
and setting $h(\mbf{x}) = 1$.
Alternatively, since 
\(
\kappa_{\text{SM}} (\mbf{x}, \mbf{y}) 
= 
\exp{\left(\frac12\|\mbf{x}\|^2 + \frac12{\|\mbf{y}\|^2}\right)} \kappa_{\text{Gauss}}(\mbf{x}, \mbf{y}),
\)
with $h(\mbf{x}) = \exp{\left( \frac12\|\mbf{x}\|^2 \right)}$ we obtain the softmax kernel.
\citet{choromanski2020rethinking} rely on the same principle, but propose an alternate construction, \textit{Positive Random Features} (PRFs), to approximate $\kappa_{\text{SM}}$, where the random feature map is defined as 
\begin{align}
    \label{eq:prf_map}
    \phi_{\mbf{\omega}}^+(\mbf{x}) 
    &\coloneqq
    \tfrac{h(\mbf{x})}{\sqrt{m}}\,
    \Bigl[\exp
    \bigl(
    \omega_1^\top \mbf{x}
    \bigr), 
    \;\dots,\;\exp
    \bigl(
    \omega_m^\top \mbf{x}
    \bigr)
    \Bigr]^\top,
\end{align}
with $h(\mbf{x}) = \exp{\left( -\frac12\|\mbf{x}\|^2 \right)}$ and again, isotropic iid projection vectors $\omega_1,\dots,\omega_m \sim \mc{N}(0, I_d)$.
Importantly, with this construction, Lemma \ref{lemma:PRFs} holds for PRFs.

\begin{lemma}[\citet{choromanski2020rethinking}]
\label{lemma:PRFs}
For $\mathbf{x}, \mathbf{y} \in \mathbb{R}^d$,
\begin{align*}
  \exp\!\bigl(\mathbf{x}^\top \mathbf{y}\bigr)
  \;=\;
  \mathbb{E}_{\mbf{\omega}\sim\mathcal{N}(0,I_d)}
  \Bigl[\,
    &\exp\!\bigl(\mbf{\omega}^\top\mathbf{x}
      \;-\;\tfrac12\|\mathbf{x}\|^2\bigr)
    \, \\
    &\exp\!\bigl(\mbf{\omega}^\top\mathbf{y}
      \;-\;\tfrac12\|\mathbf{y}\|^2\bigr)
  \Bigr].
\end{align*}
Hence, PRFs yield an unbiased approximation for the softmax kernel.
\end{lemma}

\citet{choromanski2020rethinking} argue that PRFs outperform their trigonometric counterparts, especially where kernel values are small. 
Building on this, they introduce \textit{Performer}, an efficient transformer architecture that approximates softmax attention using PRF-based random feature maps.
\citet{choromanski2020rethinking} enforce orthogonality of $\omega_1,\dots,\omega_m$ via a Gram-Schmidt orthogonalization, combining PRFs with orthogonal random features to reduce the variance of the empirical estimator.
Given a query-key pair $(\mbf{q}, \mbf{k})$, the softmax kernel in Performer is then approximated via the empirical estimator of 
\(
\bb{E}_{\omega \sim \mc{N}(0, I_d)} \left[ \phi_{\mbf{\omega}}^+(\mbf{q})^\top \phi_{\mbf{\omega}}^+(\mbf{k})\right]
\).

\begin{figure*}[t]
\centering
\includegraphics[width=0.7\textwidth]{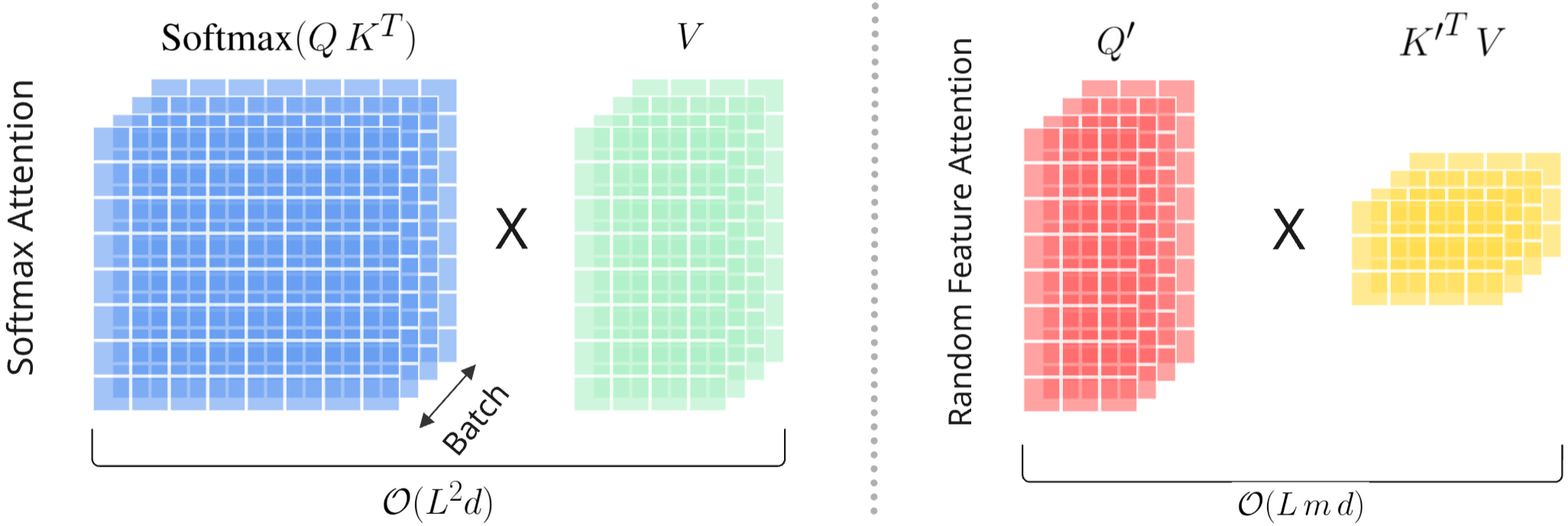}
\caption{The random feature attention replaces the softmax kernel with a linear approximation in the feature space, reducing the quadratic complexity in sequence length ($L$) to linear in sequence length times sample size ($m$).}
\label{fig:visualization}
\end{figure*}

\paragraph{Efficient Attention via Random Feature Kernels.}
The random feature map replaces the exponential kernel $\exp(QK^\top)$ with its linear approximation in the feature space $(Q^\prime (K^\prime)^\top)$ where the rows of $Q^\prime, K^\prime \in \bb{R}^{L \times m}$ are the random feature maps of the rows of $Q, K$.
Leveraging this linearity, we can approximate $\text{Att}(Q, K, V)$ without explicitly forming the $L \times L$ matrix, by first computing ${K^\prime}^\top V$ and then left multiplying it by ${Q^\prime}$. This ordering results in a time complexity of $\mc{O}(L m d)$, as illustrated in Figure \ref{fig:visualization}.
Storing $Q^\prime$ and $K^\prime$ on the other hand, requires $\mc{O}(L m)$ memory, and
the largest intermediate matrix multiplication is either
$(L \times m)\,(m \times d)$ or $(m \times L)\,(L \times d)$, 
consuming total memory on the order of
$\mc{O}\left(\max\{L m, Ld\}\right)$.
Hence, this approximation improves computation time as long as $m < L$ and memory efficiency when $\max{\left\{d, m\right\}} < L$.
Performer provides an elegant unbiased approximation of the softmax kernel via PRFs.
However, the feature sample sizes required to achieve good performance with these estimators can be so large that it dilutes the computational efficiency.
Moreover, these approximations allocate their feature budget isotropically over the input space.
For anisotropic query-key distributions, as observed in practice \citep{godey2024anisotropy}, this can waste samples on low-density directions and inflate Monte Carlo variance.
These caveats motivate our data-aligned random feature kernel, which improves approximation accuracy at smaller sample sizes by relaxing the isotropy assumption.

\section{Variance Reduction via Data‑Aware Sampling}
\label{sec:variance}

An important measure of error for random-feature kernel estimators is their Monte Carlo variance, which largely governs the performance gap with the exact exponential kernel when the number of features~$m$ is limited.
We recall that importance sampling can achieve the minimum possible variance through a proposal density that depends on the query-key distribution.
While variance-minimizing importance sampling proposals are known in principle \citep{sernau2024all}, instantiating them for PRF estimators is challenging because the optimal proposal density is typically difficult to sample from directly.
We show that the optimal proposal density is anisotropic whenever the input distribution is anisotropic, motivating anisotropy as an important dimension for improving PRF-based attention.
This variance analysis provides the theoretical motivation for DARKFormer, which realizes data-aware sampling through a learned covariance matrix, leading to the improvements we show in Section~\ref{sect:darkformer}.

Suppose the queries and keys are each iid draws from a distribution $\mc D$ on $\bb R^{d} $
with finite second moments and matching covariance matrices $\Lambda \coloneqq \operatorname{Cov}(q)
 = \operatorname{Cov}(k)$.
Let $ p_I(\omega)=\mc N(0,I_d) $
be the usual sampling density and let $ \psi(\omega) $ be any alternative density with the same support.

\begin{lemma}\label{lemma:optimal_is}
Define the PRF estimator with an importance factor $\frac{p_I(\omega_j)}{\psi(\omega_j)}$ as
\begin{align}
    \label{eq:prf_estimator_psi}
  \hat\kappa_\psi(q,k)
  :=
  \frac1m\sum_{j=1}^{m} 
  \Bigl[
      \frac{p_I(\omega_j)}{\psi(\omega_j)}
      & \exp\!\Bigl(\omega_j^{\!\top}q-\tfrac12\|q\|^2\Bigr)\, 
      \nonumber \\
      & \exp\!\Bigl(\omega_j^{\!\top}k-\tfrac12\|k\|^2\Bigr)
   \Bigr].
\end{align}
Among all finite-variance $\psi$, the expected Monte Carlo variance
\(
  \bb E_{q,k\sim\mc D}
  \!\bigl[\operatorname{Var}_{\omega}[\hat\kappa_\psi(q,k)]\bigr]
\) 
is minimized by
\[
\psi^*(\omega)\;\propto\;
p_I(\omega)\;
\sqrt{B_q(\omega)\,B_k(\omega)},
\]
where
\(
B_x(\omega)\!:=\!\bb E_x \left[ \exp\left(2\omega^{\!\top}x-\|x\|^2\right) \right]
\).
\end{lemma}

Lemma~\ref{lemma:optimal_is}, proved in Appendix~\ref{appendix:variance}, shows, for general $\psi(\omega)$, how importance sampling achieves an alignment with the input distribution, via the $B_q(\omega)B_k(\omega)$ term in the optimal $\psi^*(\omega)$, to minimize the Monte Carlo variance. 
To form a more concrete intuition for the structure of $\psi^*$ and its relationship to the isotropic baseline, consider Gaussian queries and keys, where we can characterize $\psi^*$ in closed form, assuming $\psi^*$ is normalizable, as follows (see Appendix~\ref{appendix:variance} for the proof).

\begin{theorem}\label{thm:optimal_is}
  Consider $q,k \sim \mc N(0,\Lambda)$. 
  Then the optimal sampling
  density $\psi^*$ is a centered
  Gaussian
  \(\psi^* = \mc N(0,\Sigma^*)\)
  with $\Sigma^* = (I_d+2\Lambda)(I_d-2\Lambda)^{-1}$ whenever this expression defines a valid covariance.
  Moreover:
  \begin{enumerate}
    \item $\Sigma^* \propto I_d$ if and only if $\Lambda \propto I_d$.
    \item The expected Monte Carlo variance satisfies
    \[
      \mathbb{E}_{q,k\sim\mc D}
      \bigl[\Var_\omega[\hat\kappa_{\psi^*}(q,k)]\bigr]
      \;\le\;
      \mathbb{E}_{q,k\sim\mc D}
      \bigl[\Var_\omega[\hat\kappa_{p_I}(q,k)]\bigr],
    \]
    with strict inequality for any non‑degenerate covariance
    $\Lambda \neq 0$.
  \end{enumerate}
\end{theorem}

Theorem~\ref{thm:optimal_is} implies that the optimal proposal $\psi^*$ matches the geometry of the inputs. 
$\Sigma^*$ inherits the eigenbasis of $\Lambda$, and is isotropic if and only if $\Lambda$ is isotropic.
As soon as the queries and keys have non‑trivial covariance, the optimal importance‑sampled estimator strictly reduces the expected Monte Carlo variance compared to the isotropic baseline. 
In other words, drawing projection vectors from $\mc N(0,I_d)$ is never optimal once the input distribution is non-degenerate.
However, computing this importance factor depends on the query-key distribution and involves non-trivial matrix operations.
This motivates our approach: learning a tractable, data-dependent sampling geometry that can adapt to anisotropic representations. 
We introduce DARKFormer, a random feature scheme that efficiently realizes an importance‑sampled estimator for a data-aligned kernel without explicitly forming per‑sample importance weights.


\section{Data-Aware Random Feature Kernel}
\label{sect:darkformer}

Motivated by the discussion in Section~\ref{sec:variance} on the variance optimality of adapting to input geometry via importance sampling, we introduce a \textbf{D}ata-\textbf{A}ware \textbf{R}andom-feature \textbf{K}ernel Trans\textbf{former} (\textit{DARKFormer}).
DARKFormer learns a positive semidefinite matrix $\Sigma$ (parameterized as $\Sigma=M^\top M$) that defines a data-aligned kernel and serves as the covariance of its random projections, characterizing the sampling geometry of its PRF
\footnote{For clarity and consistency of notation, we reserve $\Lambda$ for the covariance of the query--key distribution, as in Section~\ref{sec:variance}.}.
This provides a tractable mechanism for adapting random-feature attention to anisotropic query-key distributions, effectively implementing importance sampling, while bypassing explicit computation of per-sample proposal densities.
DARKFormer's data-aware approach enhances performance without extensive retraining or large feature samples. 
Moreover, we observe that it can also improve training stability. 
These improvements make the model particularly amenable to resource-constrained settings, especially for finetuning scenarios where query-key distributions are dictated by pretrained weights.

\subsection{Positive Random Features with Covariance Learning}
\label{sec:darkformer:subsec:defn}

To achieve low empirical estimation error, our proposed PRF leverages an exponential kernel of the form
\(
  \exp\bigl(q^\top \Sigma\,k\bigr),
\)
where \(\Sigma\succeq 0\) is learned to adapt to the input geometry, as is achieved by a Mahalanobis metric in the kernel's inner product space.
When $\Sigma=I_d$, this reduces to the standard softmax kernel. 
For anisotropic or correlated representations, however, the Euclidean dot-product, $q^\top k$ can be miscalibrated because it treats all coordinate directions equally and ignores cross-coordinate correlations. 
This miscalibration can be corrected by measuring similarity in a Mahalanobis geometry induced by $\Sigma$. 
When the inputs have covariance $\Lambda \succ 0$, the whitening can be achieved through $\Sigma=\Lambda^{-1}$.

\textbf{Mahalanobis Geometry for Attention.}\phantom{--}
Given $\Sigma\succ 0$, let 
\[\|x\|_{\Sigma}^{2}\coloneqq x^\top \Sigma x
~~~
\text{and}
~~~
\|x-y\|_{\Sigma}^{2}\coloneqq(x-y)^\top \Sigma (x-y),
\]
which are the Mahalanobis norm and distance, respectively. 
Then we can write 
\[
q^\top \Sigma k=\frac12(\|q\|_{\Sigma}^{2}+\|k\|_{\Sigma}^{2}-\|q-k\|_{\Sigma}^{2}).
\]
As a result, $\exp(q^\top \Sigma k)$ is, up to scaling, a Gaussian kernel in the Mahalanobis distance $\|q-k\|_{\Sigma}$. 
Equivalently, writing $\Sigma=M^\top M$ gives $q^\top \Sigma k=(Mq)^\top(Mk)$ and $\|q-k\|_{\Sigma}=\|Mq-Mk\|$.
In other words, this is the standard softmax kernel after the linear re-embedding $x\mapsto Mx$.
In particular, if the input covariance is $\Lambda\succ 0$ then choosing $M=\Lambda^{-1/2}$ (so that $\Sigma=\Lambda^{-1}$) whitens the inputs and makes $Mq$ and $Mk$ isotropic.
Thus, the Mahalanobis geometry measures distance in units of standard deviation in the principal directions, correcting the scale and correlation mismatch under anisotropic representations. 
We provide a detailed discussion and formal derivations for this in Appendix~\ref{appendix:mahalanobis}.

\paragraph{Random Feature Kernel with a Learned Geometry.}
Let \(\mbf{x} \in \bb{R}^d\) be the kernel input, and suppose \(\Sigma \in \bb{R}^{d \times d}\) is expressed as
\(
   \Sigma \;=\; M^\top M,
\)
for a learnable matrix 
\(M \in \bb{R}^{r \times d}.\)
We then write the random feature map as
\footnote{Here we write the random feature map $\phi_{\Sigma}$ as a function of $\omega$. This is a change of notation compared to Section~\ref{sect:background}, and only for ease of notation when the map involves $\Sigma$.}
\begin{align*}
    \phi_\Sigma(\mbf{x}, \tilde{\omega}) 
    &\coloneqq
    \tfrac{h(\tilde{\mbf{x}})}{\sqrt{m}}\,
    \Bigl[\exp
    \bigl(
    \tilde{\omega}_1^\top \mbf{x}
    \bigr), 
    \;\dots,\;\exp
    \bigl(
    \tilde{\omega}_m^\top \mbf{x}
    \bigr)
    \Bigr]^\top,
\end{align*}
where 
$\tilde{\omega}_1, \ldots, \tilde{\omega}_m \sim \mc{N}(0, \Sigma)$,
$h(\tilde{\mbf{x}}) = \exp{\left( -\frac12\|\tilde{\mbf{x}}\|^2 \right)}$, and $\tilde{\mbf{x}} \coloneqq M \mbf{x}$.
Given a query–key pair \(q,k \in \bb{R}^d\), the resulting PRF yields an unbiased approximation for $\exp\!\bigl(q^\top \Sigma k\bigr)$ as follows.

\begin{align}\label{eq:darkformer_kernel}
  \exp\!\bigl(\mbf{q}^\top \Sigma \mbf{k}\bigr)
  \;=\;
  \mathbb{E}_{\mbf{\tilde{\omega}}}
  \Bigl[\,
    &\exp\!\bigl(\mbf{\tilde{\omega}}^\top\mbf{q}
      \,-\,\tfrac12\mbf{q}^\top \Sigma \mbf{q}\bigr)
      \nonumber \\
    &\exp\!\bigl(\mbf{\tilde{\omega}}^\top\mbf{k}
      \,-\,\tfrac12\mbf{k}^\top \Sigma \mbf{k}\bigr)
  \Bigr].
\end{align}
The derivation of Equation \eqref{eq:darkformer_kernel} is provided in Appendix \ref{appendix:DARKFormer:kernel}.
In other words, the feature map $\phi_\Sigma(\cdot, \tilde{\omega})$ yields an exponential kernel with a learned geometry characterized by $\Sigma$. 
Learning $\Sigma$ provides a simple and flexible way to make the attention geometry data-aligned while preserving the positive random-feature structure. 
In Section~\ref{sec:darkformer:subsec:variance} we connect this covariance learning to importance sampling, which further motivates why it can reduce Monte Carlo error, in light of the discussion in Section~\ref{sec:variance}. 
This leads to closing the performance gap with exact softmax, as our experiments show, and meanwhile, it exhibits improved training stability, as observed in our experiments reported in Section \ref{sec:experiments}.

\subsection{Reducing Approximation Error via DARKFormer}
\label{sec:darkformer:subsec:variance}

We now show that DARKFormer effectively implements an importance sampling strategy. 
Section~\ref{sec:variance} shows that a fixed isotropic sampling, treating every query–key direction as equally important, is sub-optimal when the input distribution is anisotropic.
Through learning a covariance $\Sigma$ for its data-aware kernel, DARKFormer realizes an importance sampling strategy that avoids explicit computation of per-sample importance weights by coupling the sampler with the learned kernel geometry.
This in turn explains why DARKFormer can improve performance without requiring large feature samples or extensive retraining.

Let \(p_I(\omega)=\mathcal{N}(0,I_d)\) denote the isotropic sampling
distribution of the projection vectors in the PRF (Equation \eqref{eq:prf_map}) and let us denote its data-aware counterpart by \(p_\Sigma(\omega)=\mathcal{N}(0,\Sigma)\).
Viewing the change from \(p_I\) to \(p_\Sigma\) through the lens of importance sampling reveals that DARKFormer's unweighted estimator is in fact equal in expectation to an isotropic estimator whose samples are re-weighted by an importance factor.
This analogy opens the door to known results regarding the optimality of importance sampling estimators, and in particular, makes the results from Section \ref{sec:variance} applicable, which point to DARKFormer's capability in implementing a sampling scheme that can reduce estimation error. 
To see this, let us rewrite the PRF map for $\omega \sim p_{\Sigma}(\omega)$ as
\[
  \phi_{\Sigma}(x,\omega)
  \;\coloneqq\;
  \exp\!\bigl(\omega^{\top}x-\tfrac12 x^{\top} \Sigma\, x \bigr).
\]
Given a query–key pair \((q,k)\) and a feature budget of $m$ (i.e., sample size is $m$), we define the $w-$weighted empirical mean estimator for the feature kernel estimand in Equation~\eqref{eq:darkformer_kernel} as
\begin{align}
  \hat\kappa_{\Sigma}^w (q,k)
  \coloneqq\frac1m\sum_{j=1}^{m}
      w(\omega_j) \,
      \phi_{\Sigma}(q,\omega_j)\,\phi_{\Sigma}(k,\omega_j),
      \label{eq:kernel_estimator_w}
\end{align}
where the weight can depend on $\omega_j$.
Proposition \ref{prop:is} shows that sampling $\omega$ from $p_{\Sigma}(\omega)$ in the unweighted estimator is equivalent in expectation to sampling $\omega$ from the isotropic distribution $p_{I}(\omega)$ and applying importance weights (see Appendix~\ref{appendix:DARKFormer:kernel} for the proof).

\begin{proposition}
\label{prop:is}
Let $\Sigma \succ 0$ and
\(
  w_\Sigma(\omega)
  \coloneqq
  \frac{p_\Sigma(\omega)}{p_I(\omega)}.
\)
For any map \(f:\mathbb{R}^{d}\!\to\!\mathbb{R}\) with a finite mean,
\[
  \mathbb{E}_{\omega\sim p_\Sigma}\bigl[f(\omega)\bigr]
  \;=\;
  \mathbb{E}_{\omega\sim p_I}\bigl[w_\Sigma(\omega)\,f(\omega)\bigr].
\]
Consequently, for every \((q,k)\) and \(m\ge1\),
\[
  \bb{E}_{\omega_{1:m} \sim p_\Sigma} \left[ \hat\kappa_{\Sigma}^1 (q,k) \right]
  \;=\;
  \bb{E}_{\omega_{1:m} \sim p_I} \left[
  \hat\kappa_{\Sigma}^{w_{\Sigma}(\omega)} (q,k) \right]
  .
\]

\end{proposition}

This Proposition shows that drawing from $p_\Sigma$ is equivalent to re-weighted importance sampling from $p_I$. 
This reveals a key insight: 
by learning the covariance matrix $\Sigma$, DARKFormer implicitly implements importance sampling without needing to compute or store per-sample importance weights.
This is especially valuable for finetuning in resource-constrained settings, where query-key distributions are determined by pretrained weights and may require substantial retraining to conform to isotropy assumptions.

\section{Related Work}
\label{sect:related_work}

\paragraph{Random Feature Kernels.}
Kernel methods have long been valued for their ability to model complex relationships.
However, their computational cost has been a significant limitation.
Using random basis approximations \citep{rahimi2008uniform,rahimi2007random}, Random Fourier Features approximate shift-invariant kernels by mapping inputs into a randomized feature space while achieving linear complexity.
Subsequent work has characterized approximation error and improved sample efficiency via structured and orthogonal feature constructions or data-dependent sampling viewpoints such as leverage scores and kernel quadrature~\citep{sutherland2015error,le2013fastfood,yu2016orthogonal,choromanski2017unreasonable, alaoui2015fast,bach2017equivalence}.
These data-dependent views suggest that when the input distribution is anisotropic, Monte Carlo error is often dominated by a subset of directions, motivating sampling schemes that align with the geometry of the inputs rather than treating all directions equally.
These developments have enabled large-scale kernel learning in practical domains such as speech recognition \citep{may2019kernel}.

\paragraph{Efficient Transformers.}
The quadratic complexity of the attention mechanism in transformers has been a limiting factor.
Extending random feature kernels to attention, Performer \citep{choromanski2020rethinking} uses positive (orthogonal) random features \citep{choromanski2017unreasonable} to obtain an asymptotically unbiased softmax kernel estimator with linear complexity, while related linear attention variants adopt alternative normalizations \citep{katharopoulos2020transformers}.
Other approximation methods achieve efficiency through low-rank decompositions, sketching-based methods, or sparsity~\citep{wang2020linformer,xiong2021nystromformer,chen2021skyformer,lee2024proxyformer,kacham2023polysketchformer,kitaev2020reformer}.
Recent work has also studied the approximation gap of random-feature attention and proposed variance-reduction mechanisms such as control variates \citep{zheng2023efficient}.
Meanwhile, complementary approaches accelerate attention by optimizing memory access and parallelism \citep{dao2022flashattention,dao2023flashattention,shah2024flashattention,li2023distflashattn}.

\paragraph{Filling the Gap.}
Despite this progress, random-feature attention mechanisms typically sample projections from an isotropic distribution, which can exhibit high Monte Carlo variance when query--key representations are anisotropic, a common phenomenon in pretrained Transformers \citep{ethayarajh2019contextual,godey2024anisotropy}.
More broadly, data-dependent sampling can substantially reduce Monte Carlo error, but applying it to PRF-based attention is non-trivial because the optimal sampling distribution is input-dependent and costly to compute.
DARKFormer fills this gap by learning the sampling geometry through a learned covariance for positive random features, connecting to data-dependent kernel sampling \citep{alaoui2015fast,bach2017equivalence} and enabling importance-sampling behavior without explicit per-sample importance weights \citep{sernau2024all}.

\section{Experiments}
\label{sec:experiments}

To investigate DARKFormer's capabilities in empirical settings and validate the improvements we anticipate theoretically, we conduct a series of experiments on training and finetuning a Gemma-based DARKFormer.  
Replacing the exact softmax in the Gemma model \citep{team2024gemma} with the PRF-based approximation, we compare a DARKFormer model against its counterpart with isotropic PRF maps, which is a Performer-type model \citep{choromanski2020rethinking}.
While DARKFormer achieves this improvement in pretraining as well, since it is primarily intended for resource-scarce settings, our primary focus is on finetuning.
Our results demonstrate that the data-aware sampling strategy of DARKFormer translates the theoretical variance reduction benefits into practical performance improvements.
In particular, they confirm that: 
(1) DARKFormer reduces the performance gap with exact attention compared to isotropic PRF; 
(2) it achieves this improvement without requiring large feature samples or extensive training; and 
(3) it is particularly advantageous in finetuning scenarios where query-key distributions are dictated by pretrained weights.

\paragraph{Experimental Setup.}
For these experiments we use a 2-B-parameter Gemma model \citep{team2024gemma}, benchmarked on the C4 dataset \citep{raffel2020exploring} for the next-token-prediction task. 
While our primary comparison is between a DARKFormer and a Performer-type model against the exact softmax attention, we also include both learned feature kernels (LFK) and simple baselines.
In particular, for the LFK, we replace the randomly drawn projections in the PRF ($\omega$ in Equation \eqref{eq:prf_estimator_psi}) with a trainable vector. 
The LFK allows the model to freely learn the projections that yield a kernel which minimizes the task loss.
Meanwhile the simple baselines, which include a uniform random and a constant attention, benchmark all other models against a transformer-type architecture that learns around the softmax attention, highlighting how much of the change in performance is in fact due to the ability of the feature kernel used to implement an effective attention mechanism.
All comparisons are performed under identical training conditions and hyperparameters, ensuring a fair assessment.

\begin{figure}[!htb]
\centering
\includegraphics[width=0.42\textwidth]{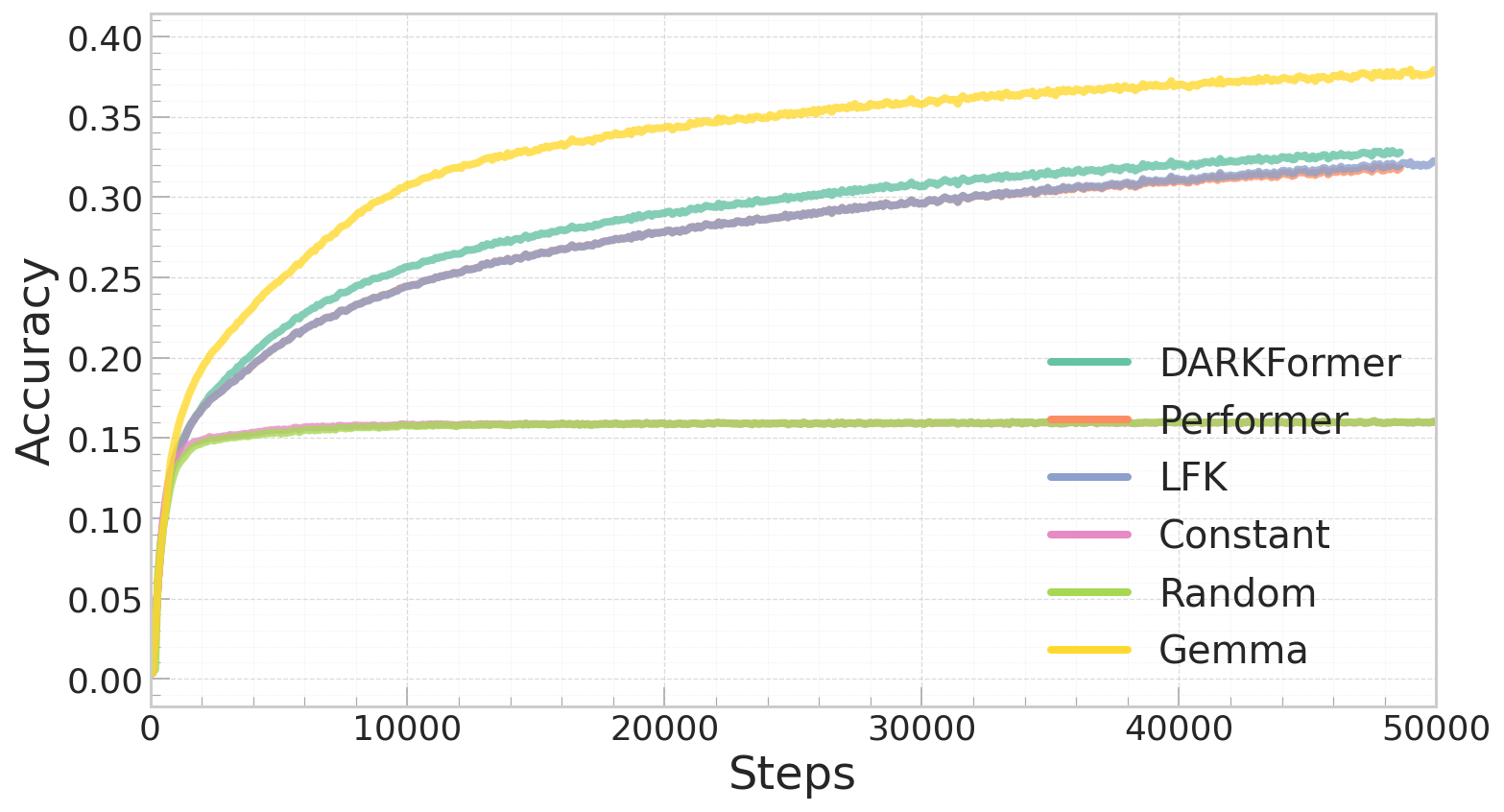}
\hspace{2em}
\includegraphics[width=0.42\textwidth]{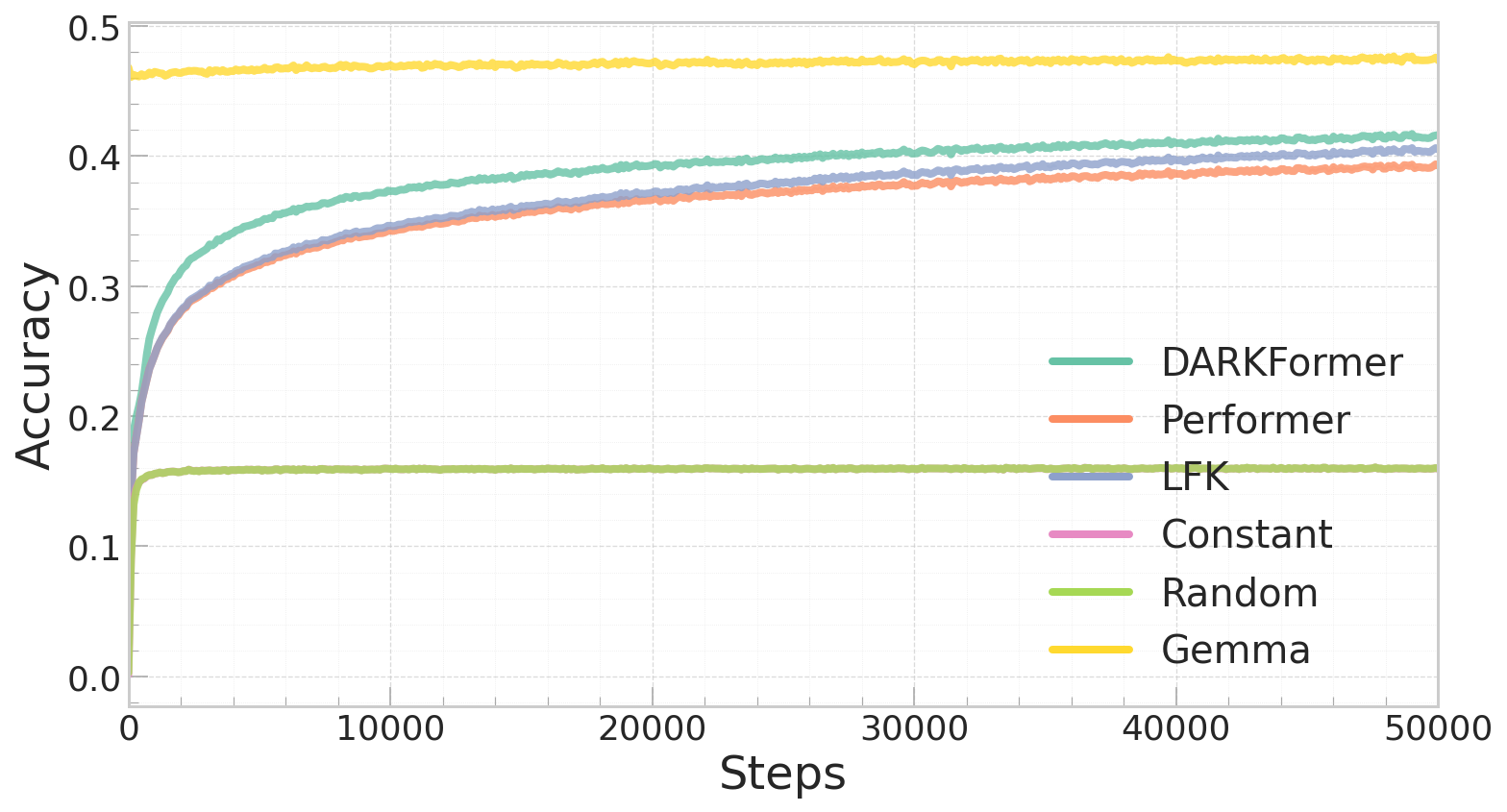}
\caption{Next token prediction accuracy during pretraining (top) and finetuning (bottom) of the Gemma-2B model with a DARKFormer (green), a Performer (orange), learned feature kernel (LFK) (blue), a random baseline (yellow), a constant baseline (lime), and an exact softmax attention. 
The DARKFormer model considerably narrows the gap between the exact softmax and the Performer-type model and also outperforms LFK, especially in finetuning.}
\label{fig:pretrain_and_finetune}
\end{figure}

\paragraph{Pretraining and Finetuning Performance.}
Starting our empirical validation with pretraining experiments, we observe that DARKFormer narrows the performance gap with the exact softmax compared with the Performer-type model.
This improvement is present throughout the pretraining as shown in Figure \ref{fig:pretrain_and_finetune}.
Turning to our main focus---finetuning---Figure \ref{fig:pretrain_and_finetune} shows that the DARKFormer model achieves significantly better next-token-prediction accuracy than the Performer model as well as all baselines, especially after the initial phase, but without extensive finetuning. 
This underscores that our proposed data-aware framework improves random-feature attention in practice, as suggested by our discussion in sections \ref{sec:variance} and \ref{sect:darkformer}, which highlight the benefits of data-aligned sampling geometries under anisotropic query-key distributions. 
Comparison with the baselines further indicates that 
(i) the observed performance is not due to the transformer learning around the PRF kernel; and (ii) the improvement is critically due to the random feature kernel used, as a learned feature kernel needs extensive finetuning to reach a performance on par with DARKFormer.

\begin{wrapfigure}[18]{r}{0.5\textwidth}
	\centering
	\captionsetup{width=.5\textwidth}
    \vspace{-.4cm}
    \includegraphics[width=.4\textwidth]{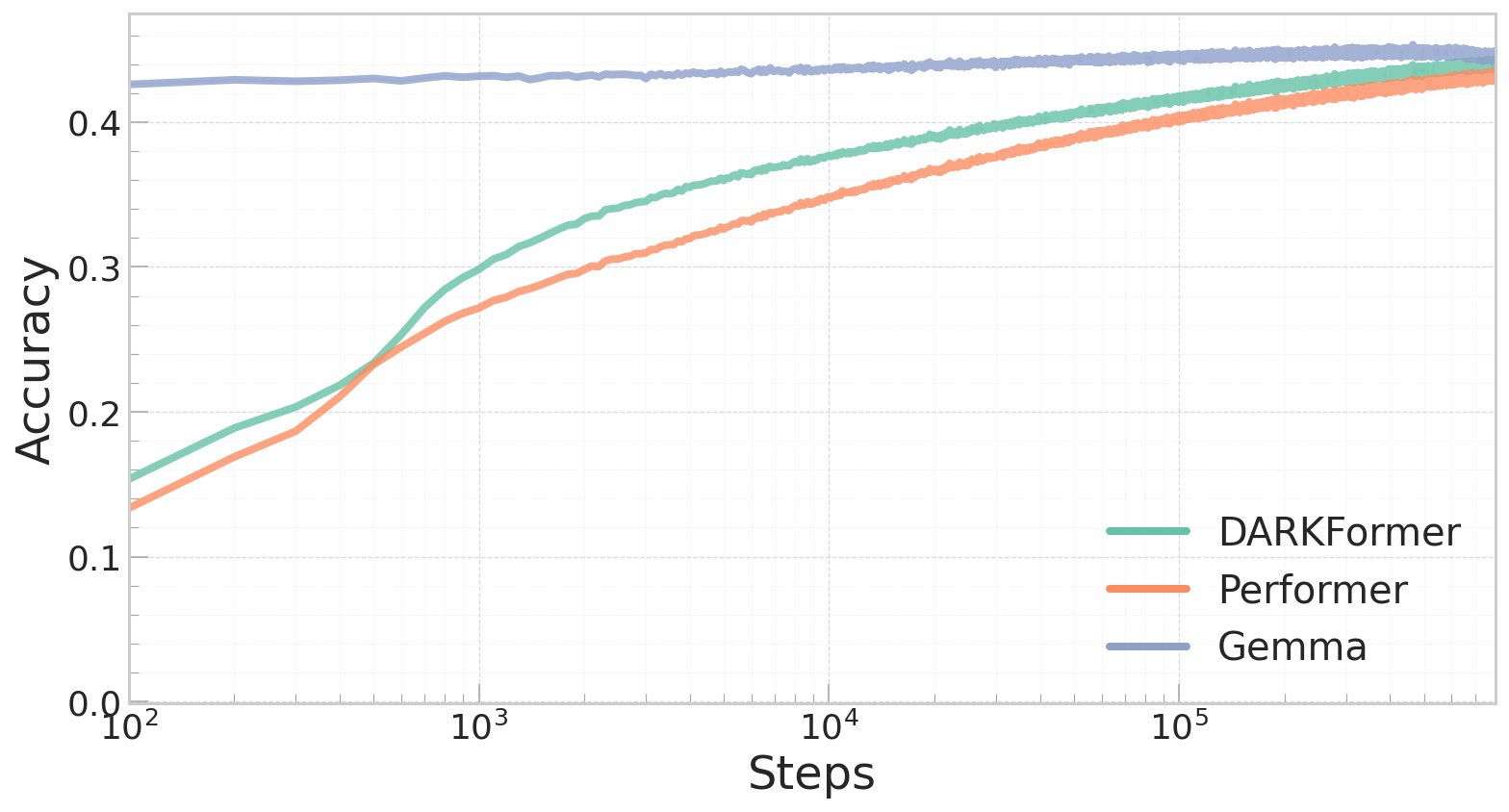}
    \caption{Next token prediction accuracy for finetuning of the Gemma-2B (blue) model with a DARKFormer (green), a Performer (orange), and an exact softmax attention over a long cycle of 650k finetuning steps. 
    Observe that DARKFormer outperforms Performer throughout training despite approximating a novel whitened kernel that is out of distribution for pretrained Gemma.
    Note that the x-axis is shown on logarithmic scale.}
    \label{fig:long_finetune}
\end{wrapfigure}

\paragraph{Efficient Finetuning with DARKFormer.}
Investigating the model's capabilities to adapt to each of the PRF kernels, we repeat our main comparison between the DARKFormer and the Performer-type models as well as Gemma with the exact softmax over extended finetuning. 
Recall that our proposed data-aligned PRF leads to an improvement in the softmax approximation when the inputs are anisotropic, which is often the case.
However, given the capabilities of transformers, they could eventually learn to produce isotropic inputs when the model is trained over a long period with isotropic PRFs.
Our observations in Figure~\ref{fig:long_finetune} confirm this: with sufficiently long finetuning the Performer-based model eventually closes much of the gap to DARKFormer, but only after many more optimization steps, incurring a significantly higher computational cost.
This highlights DARKFormer's advantage in resource-constrained scenarios, reinforcing its practical benefits for efficient finetuning.

\begin{wrapfigure}[18]{r}{0.5\textwidth}
	\centering
	\captionsetup{width=.5\textwidth}
    \vspace{-.4cm}
    \includegraphics[width=.4\textwidth]{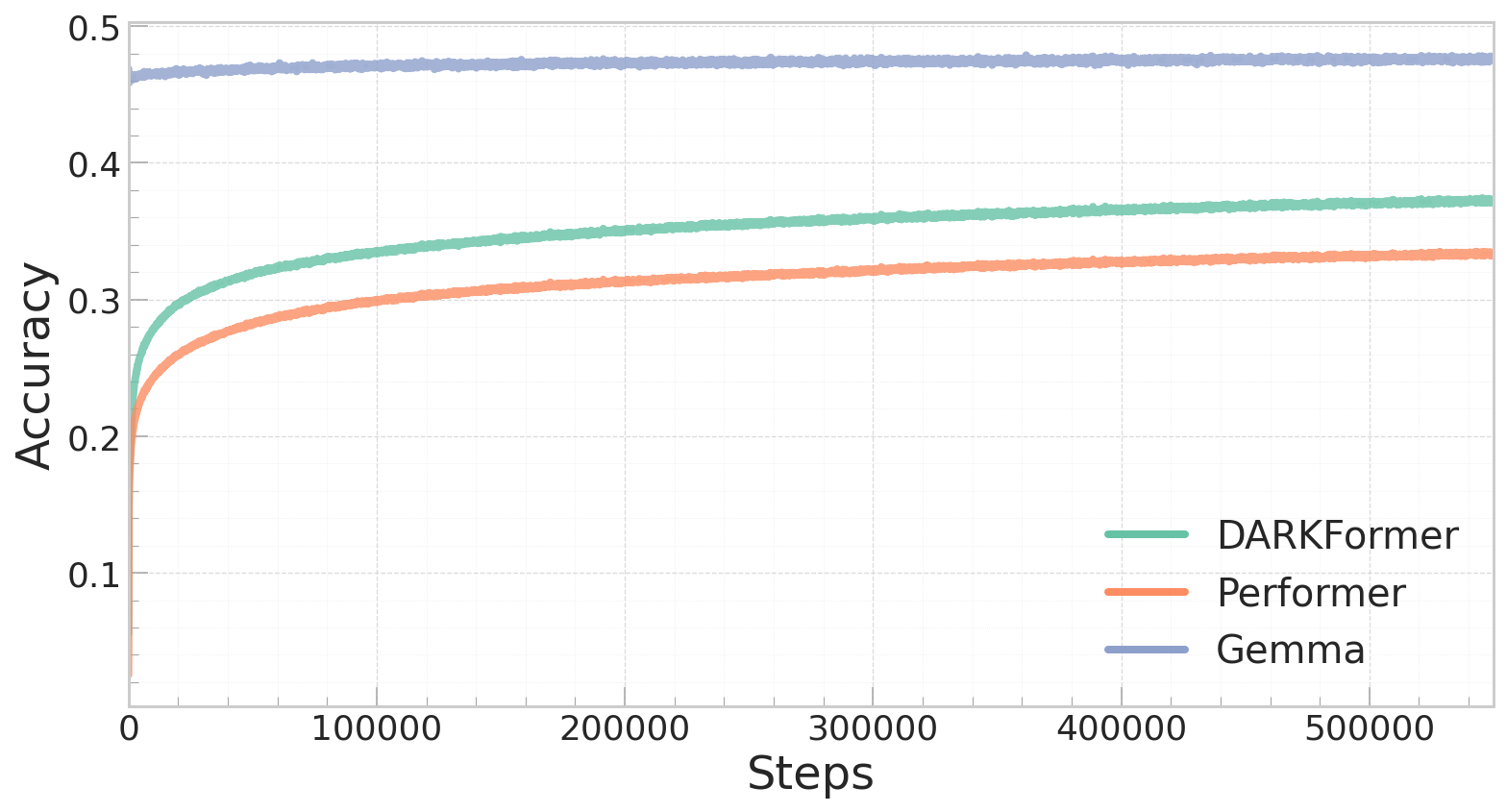}
    \caption{Next token prediction accuracy for finetuning only the q-k-v projection weights and the PRF projection covariance for the case of DARKFormer, in a Gemma-2B model with a DARKFormer (green), a Performer (orange), and an exact softmax (blue) attention over 550k finetuning steps. 
    The DARKFormer-induced improvement is even more pronounced than finetuning the full model, and does not decrease in later finetuning iterations.}
    \label{fig:freeze_finetune}
\end{wrapfigure}

\paragraph{Limited Attention Finetuning.}
In order to better disentangle effects due to the model's ability to eventually learn to produce isotropic inputs for the PRF, we investigate partial finetuning, which is also of particular interest in resource-scarce settings.
We freeze all layers of the pretrained transformer except the query, key, and value projections and, for DARKFormer, the trainable PRF covariance. 
While these projections could in principle adapt toward isotropic $q$ and $k$, freezing the rest of the network substantially limits their ability to do so.
Repeating the finetuning experiment under this setup further reveals DARKFormer's ability to improve performance, without the help of other components of the transformer.
In particular, Figure \ref{fig:freeze_finetune} shows that its performance gap with Performer is even more pronounced and does not fade over long partial finetuning, unlike the full finetuning. 
This provides further evidence on DARKFormer's capability to improve performance with limited resources.

\begin{wrapfigure}[19]{r}{0.58\textwidth}
	\centering
	\captionsetup{width=.58\textwidth}
    \includegraphics[width=.4\textwidth]{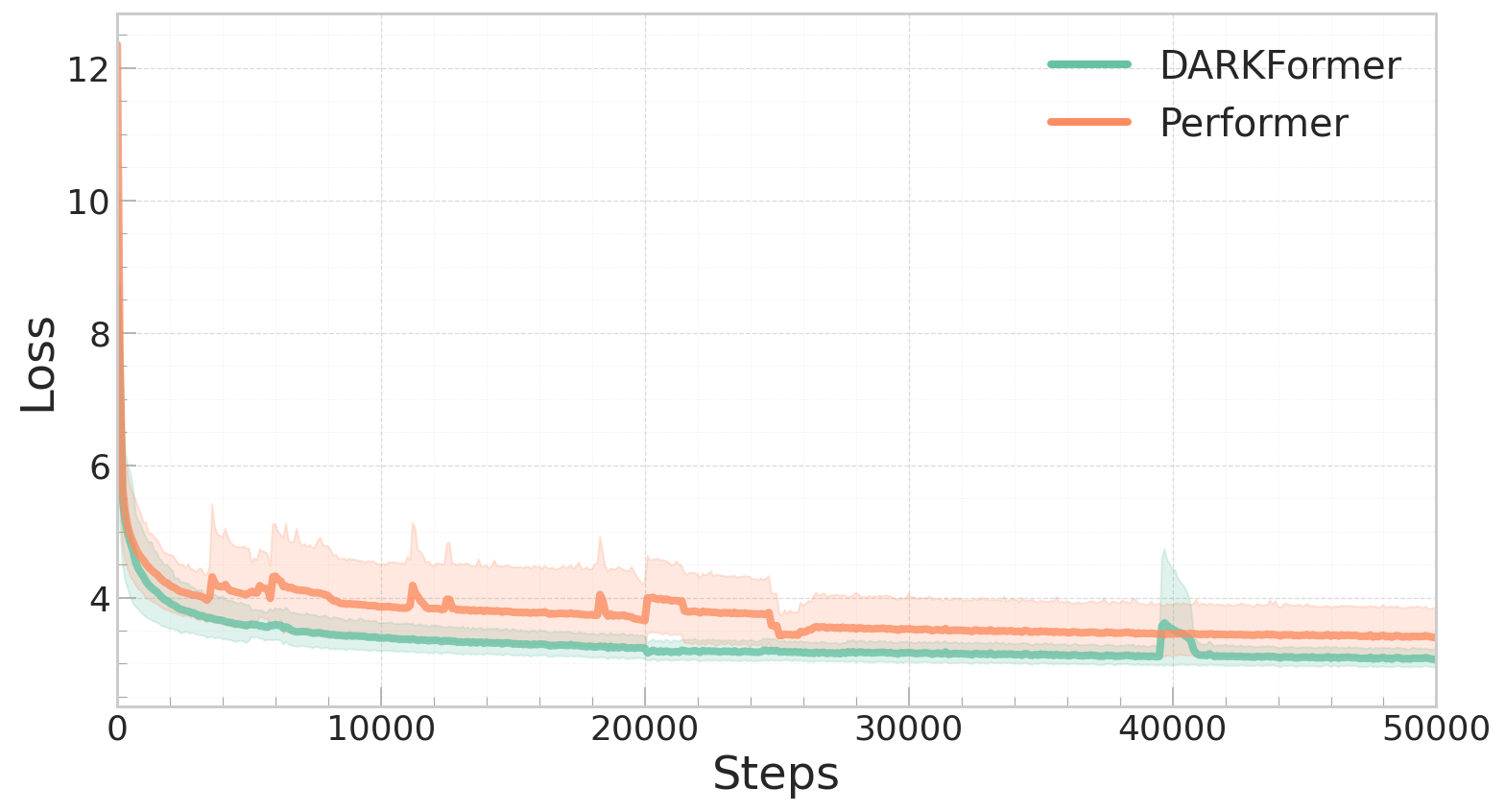}
    \caption{Loss dynamics for finetuning of the Gemma-2B model with DARKFormer (green) and a Performer (orange) with different learning rates.
    The shaded area marks the variance of loss values across seven different learning rates with the line showing the average.
    DARKFormer exhibits stable training and loss minimization throughout in all but the largest learning rate where there is a short instability phase, while Performer shows more frequent instability phases and loss spikes during its finetuning with large learning rates.
    For visualization purposes the horizontal axis is shown on logarithmic scale.}
    \label{fig:stability}
\end{wrapfigure}

\paragraph{Training Stability Across Learning Rates.}
We also evaluate training stability during finetuning by sweeping the learning rate and examining the resulting loss dynamics.
In Figure~\ref{fig:stability}, we compare DARKFormer against a Performer model under identical hyperparameters as we vary the learning rate.
Performer frequently encounters phases of numerical instability, corresponding to sharp increases in the loss value.
In contrast, DARKFormer exhibits consistently stable training dynamics across the same range of learning rates, with substantially fewer loss spikes.
Intuitively, instability under large learning rates can be triggered when stochastic updates become too noisy and lead to occasional loss-increasing steps.
From a complementary perspective, the implicit whitening of the kernel inputs implemented through the Mahalanobis inner product in DARKFormer can temper extreme dot-product magnitudes, keeping the exponential kernel in a numerically well-behaved regime.
Therefore, by aligning the random-feature sampling geometry with the input statistics, DARKFormer can not only reduce Monte Carlo variance in the attention approximation, but also make the optimization updates less erratic, which could explain the reduced frequency of loss spikes observed here.
Overall, these observations suggest that DARKFormer is more robust to the choice of learning rate, reducing sensitivity to hyperparameter tuning.
In practice, this can reduce the overhead cost of learning rate sweeps in development pipelines for training and finetuning, which is particularly valuable in resource-constrained settings.

\paragraph{Takeaways.}
Collectively, these experiments reinforce the core messages from the theoretical motivations of DARKFormer, and point to its practical advantages: 
DARKFormer's ability to reduce the performance gap with exact attention without large feature samples, its rapid adaptation during finetuning, and its enhanced effectiveness in resource-constrained scenarios all support our claims regarding the benefits of data-aware PRF kernels. 
Moreover, the stability results suggest that DARKFormer is less sensitive to learning-rate selection, reducing the need for costly extensive hyperparameter tuning.
This further reinforces DARKFormer as a practical choice for resource-constrained settings.
These findings highlight DARKFormer's contribution towards efficient transformers in scenarios where computational resources are constrained and input distributions are inherently anisotropic.

\section{Discussion}
\label{sect:discussion}

In this work, we introduced \textit{DARKFormer}, a data-aware random feature kernel for efficient transformer architectures that narrows the performance gap between exact softmax attention and positive random-feature approximations. 
By learning the covariance of the random projections, DARKFormer implements a data-aligned kernel and a corresponding importance-sampled estimator, reducing approximation error, reducing Monte Carlo variance, and enhancing training stability, with particularly strong benefits when computational resources are scarce.
Our empirical results on a Gemma model offer clear evidence for these benefits and highlight DARKFormer’s practical advantages, especially for efficient finetuning.
Looking ahead, combining data-aware random feature attention with complementary efficiency techniques and evaluating it across broader domains such as high-resolution inputs, long-context modeling, or on-device training are promising directions. 
We view DARKFormer as a step toward principled, data-aware attention mechanisms that make efficient transformers more accurate and more practical.


\bibliographystyle{unsrtnat}
\bibliography{refs}

\newpage
\appendix
\onecolumn

{\Large \textbf{Appendix}}

\section{Optimal Importance Sampling for Random Feature Kernels}
\label{appendix:variance}

In this appendix, we provide the proofs for Lemma \ref{lemma:optimal_is} and Theorem \ref{thm:optimal_is} from Section \ref{sec:variance}, which establish the theoretical foundation for variance reduction through importance sampling in random feature kernels. 
Recall that Lemma \ref{lemma:optimal_is} characterizes the optimal sampling distribution for minimizing the Monte Carlo variance of the positive random feature kernel estimator. 
Specifically, it shows that the variance is minimized by an importance factor that depends on the input distribution through the term $\sqrt{B_q(\omega)B_k(\omega)}$, as defined in the statement of the theorem. 
The subsequent Theorem \ref{thm:optimal_is} applies this result to the Gaussian case, showing that the optimal sampling distribution is isotropic if and only if the input distribution is isotropic.
These results motivate learning anisotropic sampling geometries when the input distribution is anisotropic, which informs DARKFormer’s use of a learned covariance to obtain a tractable data-aware random-feature scheme.

\paragraph{Proof of Lemma \ref{lemma:optimal_is}.}

Given \((q,k)\),  let
\(
  Z(q,k,\omega) \coloneqq
    \exp\!\bigl(\omega^{\!\top}q-\tfrac12\|q\|^2\bigr)
    \exp\!\bigl(\omega^{\!\top}k-\tfrac12\|k\|^2\bigr).
\)
By Lemma \ref{lemma:PRFs}, 
\(
  \kappa(q,k):=\bb E_{p_I}[Z]=\exp(q^{\!\top}k ).
\)
Since the summands in Equation~\eqref{eq:prf_estimator_psi} are iid, we have
\begin{align}\label{eq:var_kernel_psi}
  \Var_\omega[\hat\kappa_\psi(q,k)]
  &=
  \frac1m\Var_\psi\!\Bigl[\frac{p_I}{\psi}\,Z\Bigr]
  \;=\;
    \frac1m
    \Bigl(\bb E_\psi[(p_I/\psi)^2 Z^2]\;-\;\kappa(q,k)^2\Bigr).
\end{align}

Taking the expectation over $q,k \sim \mc D \otimes \mc D $ and discarding the
constant terms, subject to $\int\psi(\omega)\,d\omega=1$, we can write the variance reduction objective $V(\psi)$ as  
\begin{align}\label{eq:variance_obj}
  V(\psi)
  & \coloneqq \frac1m
    \int_{\bb R^{d}}
        \frac{p_I(\omega)^2}{\psi(\omega)}
        \; 
        B_q(\omega) \, 
        B_k(\omega)
        \,d\omega.
\end{align}
where
\(
B_x(\omega)\!:=\!\bb E_x \left[ \exp\left(2\omega^{\!\top}x-\|x\|^2\right) \right]
\).
Defining
\(f(\omega) \coloneqq p_I(\omega)\sqrt{B_q(\omega)B_k(\omega)}\ge0\), we can rewrite this as
\( 
V(\psi)=\tfrac1m\int f(\omega)^2/\psi(\omega)\,d\omega.
\)
For ease of notation, let
\(g_1(\omega)=f(\omega)/\sqrt{\psi(\omega)}
\) 
and
\(
g_2(\omega)=\sqrt{\psi(\omega)}
\), and note that both $g_1(\cdot)$ and $g_2(\cdot)$ are in $ L_2(\bb R^d)$, thus, we can apply Cauchy-Schwarz, which yields  
\begin{align}\label{eq:thm_is_cauchy}
  \Bigl(\textstyle\int f(\omega)\,d\omega\Bigr)^{\!2}
  =\Bigl(\int g_1(\omega) \, g_2 (\omega)\,d\omega\Bigr)^{\!2}
  \le\left(\int g_1^{2} (\omega)\,d\omega\right)\left(\int g_2^{2} (\omega)\,d\omega\right)
  =\int \frac{f^2 (\omega)}{\psi (\omega)}\,d\omega,
\end{align}
where the last equality is due to $\int\psi(\omega) \, d\omega =1$.
Note that equality holds if and only if
$g_1$ and $g_2$ are collinear, which requires $\psi\propto f$.
Now, by definition
\(f(\omega)^2=p_I(\omega)^2B_q (\omega)B_k(\omega)\),  
hence
\(
  \int f^2/\psi
  = m\,V(\psi).
\)
Plugging this in \eqref{eq:thm_is_cauchy} and rearranging gives
\[
V(\psi)\;\ge\;\frac{(\int f (\omega) \, d\omega)^2}{m},
\]
where $V(\psi)$ attains its minimum when equality holds, i.e., when $\psi\propto f$.
Thus, the optimal $\psi$ satisfies 
\(
\psi^* \propto p_I(\omega)\sqrt{B_q(\omega)B_k(\omega)}\ge0
\)
as required, completing the proof of Lemma \ref{lemma:optimal_is}. \qed

\paragraph{Proof of Theorem \ref{thm:optimal_is}.}

Let $\mc D \coloneqq \mc N(0,\Lambda)$ and write the eigen‑decomposition $\Lambda = U\,\operatorname{diag}(\lambda_1,\dots,\lambda_d)\,U^\top$ with $\lambda_i \ge 0$ and $U$ orthogonal. Since $(q,k)$ are iid draws from $\mc D$, the coordinates $q'_i,k'_i$ of $q' \coloneqq U^\top q$, $k' \coloneqq U^\top k$ are independent one‑dimensional Gaussians $\mc N(0,\lambda_i)$.

We first analyze $B_q(\omega)$ in this eigenbasis. Writing $\omega' \coloneqq U^\top \omega$ and using independence across coordinates we obtain
\begin{align*}
  B_q(\omega)
  = \mathbb E_q\!\left[\exp\bigl(2\omega^\top q - \|q\|^2\bigr)\right]
  &= \mathbb E_{q'}\!\left[\exp\bigl(2{\omega'}^\top q' - \|q'\|^2\bigr)\right]
  \\
  &= \prod_{i=1}^d
      \mathbb E_{q'_i}
        \Bigl[\exp\bigl(2\omega'_i q'_i - (q'_i)^2\bigr)\Bigr] \\
  &= \prod_{i=1}^d
        c_i\,\exp\!\bigl(\beta_i (\omega'_i)^2\bigr),
\end{align*}
where $c_i>0$ does not depend on $\omega'_i$ and $\beta_i \coloneqq \tfrac{2\lambda_i}{2\lambda_i+1} > 0$.
Hence
\[
  B_q(\omega)
  \;\propto\;
  \exp\!\Bigl(\sum_{i=1}^d \beta_i (\omega'_i)^2\Bigr),
\]
or, equivalently,  
\(
  B_q(\omega)
  \;\propto\;
  \exp\!\Bigl(\frac12 \omega^\top S \omega\Bigr),
\)
where 
\(
S 
\;\coloneqq\;
U\,\operatorname{diag}(2\beta_1,\dots,2\beta_d)\,U^\top.
\)
The same expression holds for $B_k(\omega)$ since $q$ and $k$ share the same distribution.

By Lemma~\ref{lemma:optimal_is}, the optimal density is $\psi^*(\omega) \propto p_I(\omega)\sqrt{B_q(\omega)B_k(\omega)}$.
Using $p_I(\omega) \propto \exp(-\tfrac12\|\omega\|^2)$ and $B_q=B_k$ yields
\begin{align*}
  \psi^*(\omega)
  &\propto
  \exp\!\Bigl(
    -\tfrac12\|\omega\|^2
    + \sum_{i=1}^d \beta_i (\omega'_i)^2
  \Bigr)
  \\
  &= \exp\!\Bigl(
    -\tfrac12
    \sum_{i=1}^d (1-2\beta_i)\,(\omega'_i)^2
  \Bigr),
\end{align*}
where we used $\|\omega\|^2 = \|\omega'\|^2$.
Under the technical assumption that $1-2\beta_i>0$ for all $i$, or equivalently, $\lambda_i < \frac12$ (so that $\psi^*$ is integrable), this is the unnormalized density of a centered Gaussian with covariance
\[
  \Sigma^*
  = U\,\operatorname{diag}
      \Bigl(\tfrac{1}{1-2\beta_1},\dots,\tfrac{1}{1-2\beta_d}\Bigr)
    U^\top,
\]
which proves that $\psi^* = \mc N(0,\Sigma^*)$ and establishes the first part of the theorem.

From this expression, we can see that $\Sigma^*$ clearly shares the eigenvectors of $\Lambda$. 
Moreover, each diagonal entry $1/(1-2\beta_i)$ is a scalar function of $\lambda_i$, so all eigenvalues of $\Sigma^*$ are equal if and only if all $\lambda_i$ are equal, i.e., if and only if $\Lambda \propto I_d$. 
This proves item~(1).

For item~(2), recall from Equation~\eqref{eq:variance_obj} that
\[
  V(\psi)
  \;\coloneqq\;
  \mathbb E_{q,k\sim\mc D}
  \bigl[\Var_\omega[\hat\kappa_\psi(q,k)]\bigr]
  = \frac{1}{m}\int_{\mathbb R^d}
        \frac{f(\omega)^2}{\psi(\omega)}\,d\omega,
\]
where $f(\omega) = p_I(\omega)\sqrt{B_q(\omega)B_k(\omega)}$.
The Cauchy--Schwarz inequality in~\eqref{eq:thm_is_cauchy} implies $V(\psi) \ge (\int f)^2/m$ with equality if and only if $\psi$ is proportional to $f$, i.e., $\psi=\psi^*$ after normalization.
In particular, $V(\psi^*) \le V(p_I)$, and $V(\psi^*) = V(p_I)$ can hold only if $p_I$ is proportional to $f$.

In the Gaussian setting above, $B_q(\omega)$ is non‑constant whenever $\Lambda\neq 0$, so $f(\omega)\propto p_I(\omega)$ can occur only in the degenerate case $\Lambda = 0$. 
Consequently, for any non‑degenerate covariance $\Lambda\neq 0$ we have $V(\psi^*) < V(p_I)$, proving item~(2).
\qed

\section{Data-aware Positive Random Feature Kernel: Derivations}
\label{appendix:DARKFormer:kernel}

This appendix provides the derivations of the results on DARKFormer's PRF kernel, presented in Section \ref{sect:darkformer}. 
We first derive the expression for the exponential kernel with learned covariance given in Equation \eqref{eq:darkformer_kernel}, which forms the core of our data-aware random feature kernel. 
We then prove Proposition \ref{prop:is}, which establishes the equivalence between DARKFormer's covariance learning and importance sampling---a critical insight that allows our method to implicitly implement importance sampling.

\paragraph{Derivation of Equation \eqref{eq:darkformer_kernel}.}

Here we derive the expression for DARKFormer's data-aware kernel as presented in Equation \eqref{eq:darkformer_kernel}, showing how a learned covariance matrix $\Sigma$ can produce an unbiased approximation of the exponential kernel $\exp(q^\top \Sigma k)$ as discussed in Section \ref{sec:darkformer:subsec:defn}.
Let $\Sigma\in\bb{R}^{d\times d}$ be positive–semidefinite and write
$\Sigma = M^\top M$ with $M\in\bb{R}^{r\times d}$ (\,$r\le d$\,).
We aim to show, for any $\mbf{q},\mbf{k}\in\bb{R}^{d}$,
\begin{equation}\label{eq:lemma-covPRF}
  \exp\!\bigl(\mbf{q}^{\top}\Sigma\,\mbf{k}\bigr)
  \;=\;
  \bb{E}_{\tilde\omega\sim\mathcal N(0,\Sigma)}
  \Bigl[
     \exp\!\bigl(
         \tilde\omega^{\top}\mbf{q} - \tfrac12\,\mbf{q}^{\top}\Sigma \mbf{q}
       \bigr)\,
     \exp\!\bigl(
         \tilde\omega^{\top}\mbf{k} - \tfrac12\,\mbf{k}^{\top}\Sigma \mbf{k}
       \bigr)
  \Bigr].
\end{equation}

Let $\tilde q := Mq,\;\tilde k := Mk\in\bb{R}^{r}$.
Applying Lemma \ref{lemma:PRFs} in \(\bb{R}^{r}\) with \(w\sim\mathcal N(0,I_r)\) gives
\begin{align}\label{eq:ref_kernel}
  \exp(\tilde q^{\top}\tilde k)
  &= 
  \bb{E}_{w\sim\mathcal N(0,I_r)}
    \left[
      \exp{\left(w^{\top}\tilde q-\frac12\|\tilde q\|^{2}\right)}\;
      \exp{\left(\,w^{\top}\tilde k-\frac12\|\tilde k\|^{2}\right)}
    \right].
\end{align}
Replacing $w$ with $\tilde\omega := M^{\!\top}w$ in the expectation,
the expectation will be taken with respect to $\tilde\omega\sim \mc N(0,\Sigma)$.
Moreover
$w^{\top}\tilde q = w^{\top}Mq = (M^{\!\top}w)^{\top}q=\tilde\omega^{\top}q$
and similarly $w^{\top}\tilde k=\tilde\omega^{\top}k$,
while $\|\tilde q\|^{2}=q^{\top}\Sigma q$ and
$\|\tilde k\|^{2}=k^{\top}\Sigma k$.
Substituting these into Equation \eqref{eq:ref_kernel} yields Equation \eqref{eq:darkformer_kernel}.

\paragraph{Proof of Proposition \ref{prop:is}.}
Here we prove Proposition \ref{prop:is}, which establishes that sampling projection vectors from a non-isotropic distribution $p_\Sigma(\omega)$ is equivalent to sampling from the isotropic distribution $p_I(\omega)$ and applying importance weights.
For any integrable $f$,
\[
  \bb{E}_{\omega\sim p_\Sigma}[f(\omega)]
  =\!\int f(\omega)\,p_\Sigma(\omega)d\omega
  =\!\int f(\omega)\,w_\Sigma(\omega)\,p_I(\omega)d\omega
  =\bb{E}_{\omega\sim p_I}[w_\Sigma(\omega)f(\omega)] .
\]
Choosing $f(\omega)=\phi_\Sigma(q,\omega)\,\phi_\Sigma(k,\omega)$ gives
\[
\bb{E}_{\omega\sim p_\Sigma}\!\big[\phi_\Sigma(q,\omega)\phi_\Sigma(k,\omega)\big]
=
\bb{E}_{\omega\sim p_I}\!\big[w_\Sigma(\omega)\phi_\Sigma(q,\omega)\phi_\Sigma(k,\omega)\big],
\]
and hence, for the empirical average with iid\ sampling we have
\(
\bb{E}_{\omega_{1:m}\sim p_\Sigma}\!\left[\hat\kappa_{\Sigma}^1(q,k)\right]
=
\bb{E}_{\omega_{1:m}\sim p_I}\!\left[\hat\kappa_{\Sigma}^{w_\Sigma}(q,k)\right].
\)
\qed

\section{Attention with Mahalanobis Geometry}
\label{appendix:unbiased}
\label{appendix:mahalanobis}
\label{appendix:reference}
\label{sec:darkformer:subsec:reference}

In Section~\ref{sect:darkformer} we introduce a data-aware kernel of the form $\exp(q^\top \Sigma k)$. 
This appendix provides additional details on this kernel. 
In particular, we explain how the $\Sigma$-inner product measures similarity through Mahalanobis distance and show how choosing $\Sigma$ as an inverse covariance whitens the representation space. 
This formalizes why replacing $q^\top k$ by $q^\top \Sigma k$ corrects both scale and correlation mismatch under anisotropic representations.

\paragraph{Attention with Mahalanobis Geometry.}
For $\Sigma\succ 0$, define the Mahalanobis norm $\|x\|_{\Sigma}^{2}\coloneqq x^\top \Sigma x$ and the induced distance $\|x-y\|_{\Sigma}^{2}=(x-y)^\top \Sigma (x-y)$. 
Note that
\(
x^\top \Sigma y
=
\frac12\bigl(\|x\|_{\Sigma}^{2}+\|y\|_{\Sigma}^{2}-\|x-y\|_{\Sigma}^{2}\bigr).
\)
Consequently,
\[
\exp(x^\top \Sigma y)
=
\exp\!\Bigl(\tfrac12\|x\|_{\Sigma}^{2}\Bigr)\,
\exp\!\Bigl(\tfrac12\|y\|_{\Sigma}^{2}\Bigr)\,
\exp\!\Bigl(-\tfrac12\|x-y\|_{\Sigma}^{2}\Bigr).
\]
Hence, replacing $q^\top k$ by $q^\top \Sigma k$ corresponds to measuring similarity in a Mahalanobis geometry. 
In particular, $\exp(q^\top \Sigma k)$ is, up to scaling, a Gaussian kernel in the Mahalanobis distance $\|q-k\|_{\Sigma}$. 
This directly addresses issues that arise under anisotropic representations, since $\Sigma$ can downweight high-variance directions and incorporate correlations across coordinates. 

\begin{proposition}
\label{prop:mahalanobis}
Given $\Sigma\succ 0$, let $M$ be the matrix satisfying $\Sigma=M^\top M$ and suppose $r=d$. 
Then for all $q,k\in\bb{R}^d$ we have $q^\top \Sigma k=(Mq)^\top(Mk)$ and $\|q-k\|_{\Sigma}^{2}=\|Mq-Mk\|^{2}$. 
If the representation covariance is $\Lambda=\mathrm{Cov}(q)=\mathrm{Cov}(k)\succ 0$ and we choose $\Sigma=\Lambda^{-1}$ with $M=\Lambda^{-1/2}$, then $\mathrm{Cov}(Mq)=I_d$ and
\(
\|q-k\|_{\Lambda^{-1}}^{2}
=
\sum_{i=1}^d \delta_i^2/\lambda_i,
\)
where $\Lambda=U\mathrm{diag}(\lambda_1,\dots,\lambda_d)U^\top$ and $\delta=U^\top(q-k)$. 
\end{proposition}

\paragraph{Proof.}
For any $q,k\in\bb{R}^d$ we have
\(
q^\top \Sigma k
=
q^\top M^\top M k
=
(Mq)^\top(Mk).
\)
Similarly,
\[
\|q-k\|_{\Sigma}^{2}
=
(q-k)^\top \Sigma (q-k)
=
(q-k)^\top M^\top M (q-k)
=
\|M(q-k)\|^{2}
=
\|Mq-Mk\|^{2}.
\]
Now assume $\Lambda=\mathrm{Cov}(q)=\mathrm{Cov}(k)\succ 0$ and choose $\Sigma=\Lambda^{-1}$ with $M=\Lambda^{-1/2}$. 
Then
\[
\mathrm{Cov}(Mq)
=
M\,\mathrm{Cov}(q)\,M^\top
=
\Lambda^{-1/2}\,\Lambda\,\Lambda^{-1/2}
=
I_d.
\]
Now write $\Lambda=U\mathrm{diag}(\lambda_1,\dots,\lambda_d)U^\top$ with $U$ orthogonal and $\lambda_i>0$. 
Then $\Lambda^{-1}=U\mathrm{diag}(\lambda_1^{-1},\dots,\lambda_d^{-1})U^\top$, and 
with $\delta\coloneqq U^\top(q-k)$, we obtain
\[
\|q-k\|_{\Lambda^{-1}}^{2}
=
(q-k)^\top \Lambda^{-1}(q-k)
=
\delta^\top \mathrm{diag}(\lambda_1^{-1},\dots,\lambda_d^{-1})\delta
=
\sum_{i=1}^d \delta_i^2/\lambda_i,
\]
as claimed. 
\qed

The identity $\|q-k\|_{\Lambda^{-1}}^{2}=\sum_i (\delta_i/\sqrt{\lambda_i})^{2}$ shows that the Mahalanobis distance measures displacements in units of principal standard deviations. 
Thus, directions with large variance are naturally downweighted and correlations are removed by whitening. 
In the attention kernel, this corresponds to applying the standard dot-product softmax kernel to the re-embedded coordinates $\tilde q=Mq$ and $\tilde k=Mk$. 
This motivates learning $\Sigma$ as a tractable way to adapt the attention geometry to anisotropic inputs, rather than relying on the transformer to reshape representations toward isotropy.


\end{document}